\documentclass{article}

\usepackage[preprint]{neurips_2021}

\usepackage{natbib} 

\usepackage{mathtools} 
\usepackage{booktabs} 

\usepackage[utf8]{inputenc} 
\usepackage[T1]{fontenc}    
\usepackage{hyperref}       
\usepackage{url}            
\usepackage{booktabs}       
\usepackage{amsfonts}       
\usepackage{nicefrac}       
\usepackage{microtype}      
\usepackage{xcolor}         

\renewcommand{\vec}[1]{\boldsymbol{#1}}     
\newcommand{\pvec}[1]{#1}     
\newcommand{\mat}[1]{\mathbf{#1}}           
\newcommand{\R}{\mathbb{R}}

\usepackage[ruled,vlined]{algorithm2e}
\newlength\mylen

\SetKwInput{KwInput}{Input} 
\SetKwInput{KwParams}{Hyper-parameters}

\usepackage[american]{babel}
\usepackage[mathscr]{euscript}

\usepackage{mathtools} 
\usepackage{booktabs} 
\usepackage{tikz} 
\usepackage{amssymb,bm}
\usepackage[ruled,vlined]{algorithm2e}
\usepackage{subcaption}

\renewcommand{\vec}[1]{\boldsymbol{#1}}     

\title{Learning to Optimize with Dynamic Mode Decomposition}

\author{%
  Petr Šimánek\\
    Faculty of Information Technology, Czech Technical University in Prague\\
    GoodAI\\
  \texttt{petr.simanek@fit.cvut.cz} \\
\And

  Daniel Vašata\\
    Faculty of Information Technology\\
    Czech Technical University in Prague\\
  \texttt{daniel.vasata@fit.cvut.cz} \\
  \And
    Pavel Kordík\\
    Faculty of Information Technology\\
    Czech Technical University in Prague\\
  \texttt{kordikp@fit.cvut.cz} \\

}

\begin{document}

\maketitle

\begin{abstract}
  Designing faster optimization algorithms is of ever-growing interest. In recent years, learning to learn methods that learn how to optimize demonstrated very encouraging results. Current approaches usually do not effectively include the dynamics of the optimization process during training. They either omit it entirely or only implicitly assume the dynamics of an isolated parameter. In this paper, we show how to utilize the dynamic mode decomposition method for extracting informative features about optimization dynamics. By employing those features, we show that our learned optimizer generalizes much better to unseen optimization problems in short. The improved generalization is illustrated on multiple tasks where training the optimizer on one neural network generalizes to different architectures and distinct datasets. 
\end{abstract}

\section{Introduction}\label{sec:intro}

Methods called \emph{learning to learn} (L2L) or meta-learning are designed to speed up the learning process by leveraging knowledge obtained in past learning sessions. Typically, there are two learning loops present, where the inner loop is responsible for rapid learning on a particular task while the outer loop is a slower learning process over a set of similar tasks or past learning sessions. L2L methods are currently intensively studied \citep{review} with many different applications, for example, one-shot or few-shot learning. A very thorough review of the historical development of L2L methods can be found in \citep{schmidhuber}.

One particular application or even a subfield of L2L algorithms is \emph{learning to optimize} (L2O) focusing on improving the performance of optimization methods in a meta-learning fashion. There is a substantial body of literature devoted to learning to optimize with available gradient information. The proposed methods try to learn the "gradient descent-like" method to optimize the parameters (weights and biases) of neural networks. The methods range from learning adaptive step-size (learning rate) to learning the whole gradient descent algorithm \citep{Andrychowicz, rnnprop, metz2}. The resulting methods tend to suffer from limited generalization, i.e. fail when applied to tasks very different from the training task. 
These methods often use a recurrent neural network that learns how to update parameters in the inner loop. This recurrent neural network is trained by standard methods like Adam \citep{Adam} in the outer loop. 
Learning to optimize without gradient information is also an important field of study \citep{l2lw} with applications in e.g. optimization of processes in chemical reactions \citep{chemopt}. These approaches are quite similar to \citep{Andrychowicz} as they use LSTM \citep{LSTM} as an optimizer, but they cast the optimization problem in a reinforcement learning manner.
A more general and ambitious approach is proposed in Badger \citep{badger} and VS-ML \citep{vsml}. The goal is to learn a very general rule that optimizes any function and which generalizes very well outside the training tasks.

In this paper, we closely follow the work of \citep{Andrychowicz}. The authors propose a method where one LSTM recurrent neural network called optimizer (or expert in terms coined in \citep{badger}) is responsible for updating one parameter in the optimized task (called optimizee). This results in thousands of LSTMs when optimizing even small neural networks. To make this method practically useful and the training possible, LSTM weights are shared between all the LSTMs. This results in a favorable ratio between the number of parameters of the optimizer and the optimizee \citep{badger, vsml}. However, there is no communication between optimizers, which might hamper the quality of the overall optimization. 

In this work, we introduce an optimizer that benefits from knowing the dynamics of each optimized parameter and also the dynamics of the optimizee (i.e. loss in case of neural networks training). This is achieved by calculating the eigenvalues of the Dynamic Mode Decomposition (DMD) and using them as a further input to the optimizer.
The dynamic mode decomposition method \citep{Schmid} is currently intensively studied and applied in many areas as e.g. fluid dynamics or stock trading. The DMD is a method similar to PCA and Fourier transformation but allows us to easily extract more information about the immediate dynamics and also provides dimensionality reduction. The DMD method is closely connected to the Koopman operator, see e.g. \citep{tu2014} and related methods such as EDMD \citep{EDMD}. Various applications of the theory behind Koopman operator in the context of artificial neural networks were studied in \citep{Mezic} and \citep{redman}. The Koopman operator proposes a way how to transform a nonlinear dynamical system into a linear (but infinite-dimensional) system. This transformation can further lead us to derive Koopman eigenvalues and modes which describe the dynamics. It was proven by \citet{tu2014} that DMD approximates the Koopman operator under some conditions. We observe that the aggregate information about the dynamics of all optimized parameters of the optimizee is beneficial for the training of the optimizee by the learned optimizer. Our experiments also indicate that the method works well for different optimization tasks than it was trained for.
\section{Related Work}
There are currently many interesting ways how to increase the generality and performance of a learned optimizer.
\citet{rnnprop} propose a few tricks that allow optimization for longer horizons, namely random scaling of the gradients. Apart from this training trick they propose a method called \emph{rnnProp} which uses a more complicated update rule similar to Adam. This approach allows the the optimizer to train for long horizons, but it can be argued that the optimizer learns only to imitate the the background methods (Adam). Once the gradients are normalized and also the structure of Adam update is prescribed, the optimizer can learn identity function and produce a good results (e.g. it learns only learning rate). We believe that such approach will in principle be only a marginally better than hand-designed methods. 

A much larger change in the RNN optimizer architecture proposed in \citep{l2o-scale} is called \emph{L2O-Scale}. Hierarchical RNN with minimized per-parameter overhead and also mirroring known optimizers improves performance of learning to optimize. The authors suggest also meta-training on a set of optimization tasks.  

Recently, \citet{ll2o} demonstrated that advanced training techniques substantially improve the trained optimizers. Even the basic L2O model outperforms some recent sophisticated models, e.g. rnnProp and L2O-Scale mentioned before. They propose a progressive training scheme and introduce off-policy imitation learning by imitating the behavior of a successful optimizer.     

Most of the learning to optimize approaches are trained on one task or on a small set of tasks. \citet{metz} apply a radically different approach and train the optimizer on thousands of optimization tasks. Apart from this approach, they propose a hierarchical network that can use additional information specific to the task, e.g. validation loss. 

We believe that the method introduced in this paper is very general and should be useful for any of these approaches.


\section{Preliminary Methods}
Let us start with explaining the learning to optimize method that represents an initial model for our considerations. Then the dynamic mode decomposition follows as an important tool how to improve the performance of the L2O.

\subsection{Learning to Optimize}\label{section:l2o}
The primary task under consideration is to minimize a function \(L(\pvec\theta)\) with respect to its vector of parameters $\pvec\theta$. Let us now follow \citet{Andrychowicz}.
The key idea of the learning to optimize method is to train a recurrent neural network \(M\) parametrized by $\pvec \phi$
that acts as an optimizer suggesting updates of parameters $\pvec\theta_t \mapsto \pvec\theta_{t+1}$ aiming the training to converge to some local minimum of $L$. The network \(M\) is then called the optimizer (or meta-learner) and \(L(\pvec\theta)\) the optimizee.

The task of the optimizer is to suggest at every training step $t$ an update $\pvec g_t$ of parameters $\pvec\theta_t$ of the optimizee based on the gradient $\nabla_{\pvec\theta} L(\pvec\theta_t)$ and on its own hidden states $\pvec h_t$, which is also updated in each step:
\begin{equation}\label{eq:L2O-train}
    \pvec\theta_{t+1}=\pvec\theta_t + \pvec g_t,
\end{equation}
where
\begin{equation}\label{eq:L2O-update}
    [\pvec g_t, \pvec h_{t + 1}] = M\big(\nabla_{\pvec\theta} L(\pvec\theta_t), \pvec h_t, \pvec\phi\big).
\end{equation}
The parameters of the optimizee are updated by the optimizer at every training step.
The parameters $\pvec \phi$ of the optimizer $M$ are learned by stochastic gradient descent and 
updated every $u$-th training step where a hyper-parameter $u$ is called the unroll. 
The loss of the optimizer is the expectation of 
the weighted unrolled part of the optimizee training trajectory,
\begin{equation}\label{eq:optimizer-objective}
    \mathscr L(\pvec \phi) = \mathbb{E}_L\sum_{\tau = 1}^{u} w_\tau L(\pvec \theta_{\tau + j u - 1}),
\end{equation}
where $\mathbb E_L$ is the expectation with respect to some distribution of optimizee functions $L$, $w_\tau$ are weights that are typically set to $1$, and $j u$ is the initial training step of the last unrolled part that therefore corresponds to training steps $t = ju, ju + 1, \dotsc, (j+1)u - 1$. In practice one often has only one optimizee function $L$ so the expectation in \eqref{eq:optimizer-objective} is removed.

In addition to updates of $\pvec \phi$ along the optimizee training, there is also an outer loop where 
the whole training of the optimizee is restarted while parameters $\pvec \phi$ continue to learn.

  \begin{figure}
    \begin{subfigure}[b]{0.45\linewidth}
      \includegraphics[width=1\linewidth]{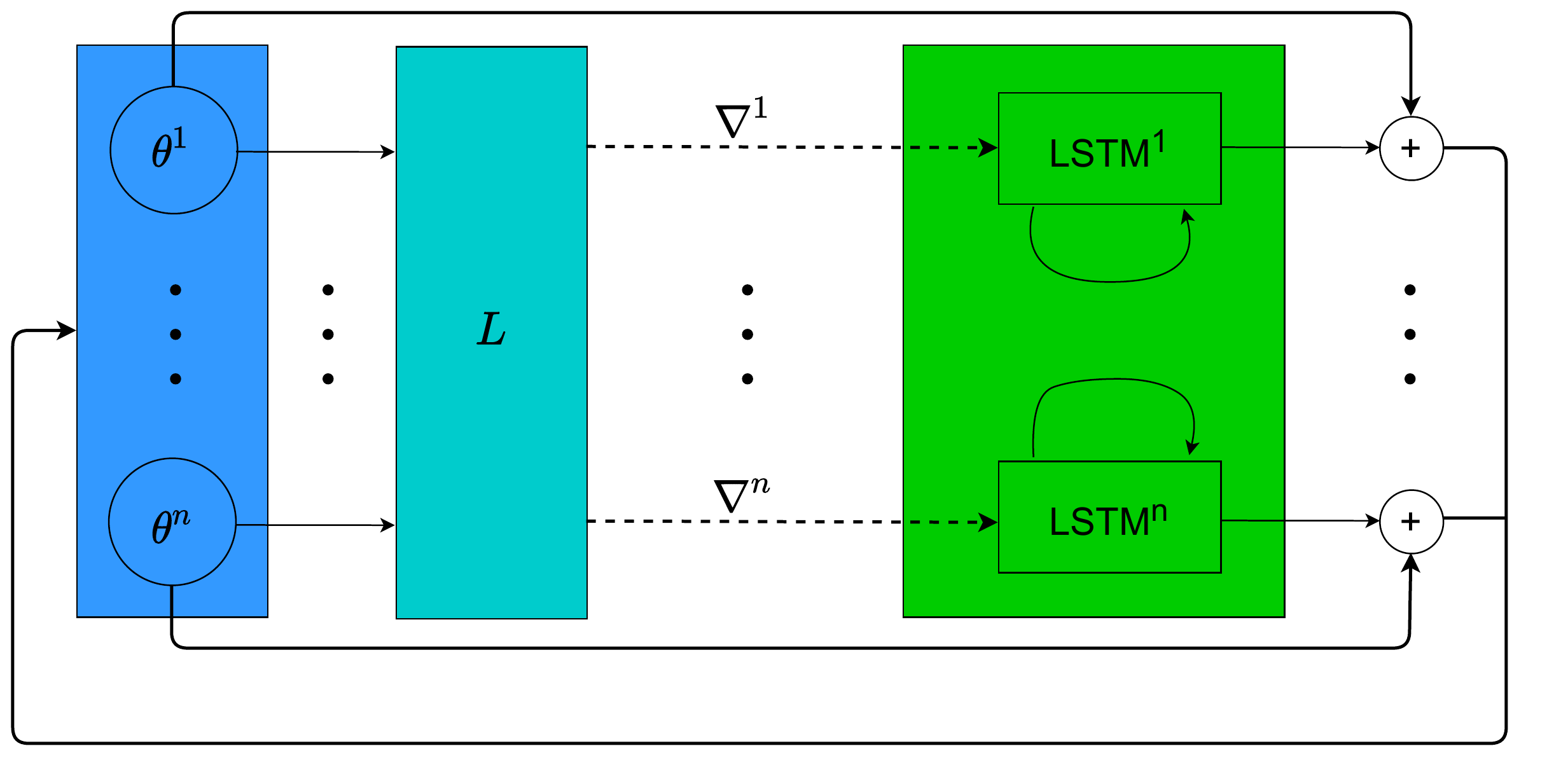}
      \caption{Original L2O.}\label{fig:archa}
    \end{subfigure}
    \begin{subfigure}[b]{0.45\linewidth}
      \includegraphics[width=1\linewidth]{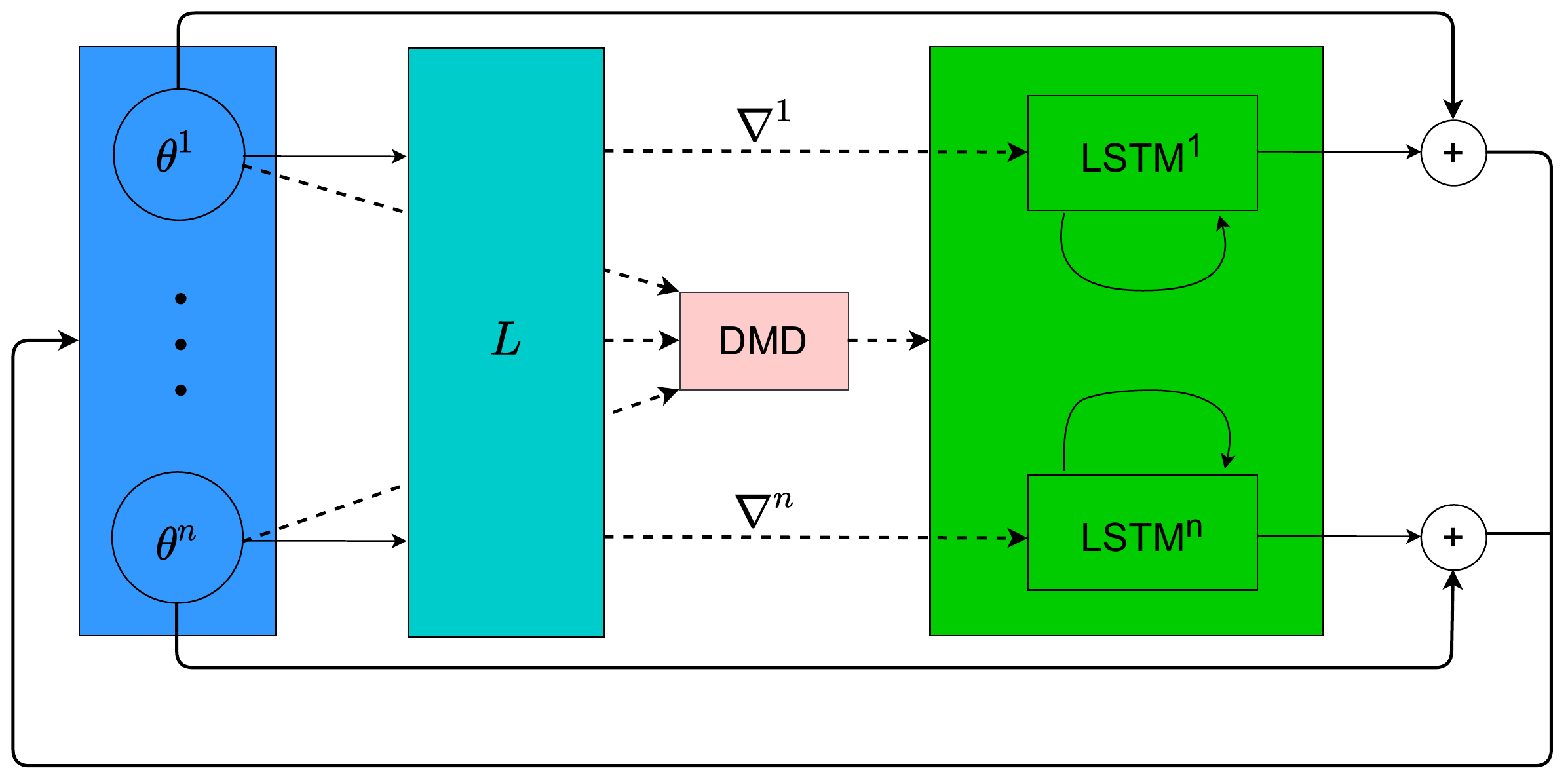}
    \caption{L2O-DMD Optimizer.}\label{fig:archb}
    \end{subfigure}
    \caption{One Step of an L2O Optimizer}
  \end{figure}

For further algorithmic details and pre-processing we refer the reader to \citep{Andrychowicz}.

In practice when there are several thousand or more parameters in $\theta$, the above application of a general recurrent neural network is an almost impossible task. 
The authors of \citep{Andrychowicz} overcome this issue by implementing the update rule in a coordinate-wise manner using a two-layer LSTM network with shared parameters. This means that $M$ is a small network with multiple instances operating on each parameter of the optimizee separately while sharing its parameters $\phi$ across all instances. We will refer to this approach as to the original L2O. The visualisation of one optimization step is presented in Figure \ref{fig:archa}.

\subsection{Dynamic Mode Decomposition}
In this part, we briefly summarize the dynamic mode decomposition method (DMD). 
For more details, we refer the reader to e.g. \cite{Schmid,tu2014}.
The DMD is typically concerned with data from a dynamical system 
 $   \vec{x}_{k+1}= G(\vec{x}_k),$
where \(\vec{x}_k \in \R^n\) is a state vector of the dynamical system at time \(k\) (called snapshot) and \(G\) represents the dynamics (evolution) of the system which can be nonlinear. 

In the exact DMD one constructs a matrix \(\mat{A} \in \R^{n,n}\) generating an approximate locally linear dynamical system 
 $   \vec{x}_{k+1} \approx \mat{A}\vec{x}_k$
that fits the measured trajectory in a least-squares style.
Hence, one wants to minimize the overall approximation error given by the sum of Euclidean norms:
\begin{equation}\label{eq:dmd-to-minimize}
    \sum_{k = 1}^{m-1}\lVert\vec{x}_{k+1} - \mat{A}\vec{x}_k\rVert_2^2.
\end{equation}
To write it in the matrix form let us arrange the snapshots into two matrices 
composed of column vectors in the following way:
\begin{equation*}
    \mat{X} = 
    \begin{pmatrix}
        | & | &  & | \\
        \vec{x}_0 & \vec{x}_1 & \cdots & \vec{x}_{m-1} \\
        | & | &  & |  
    \end{pmatrix},
    \mat{Y} = 
    \begin{pmatrix}
        | & | &  & | \\
        \vec{x}_1 & \vec{x}_2 & \cdots & \vec{x}_{m} \\
        | & | &  & |  
    \end{pmatrix}.
\end{equation*}
Under this notation the minimization of \eqref{eq:dmd-to-minimize} is equivalent to the minimization of the Frobenius norm of the matrix $\mat Y - \mat A \mat X$.
The optimal \(\mat{A}\) which minimizes the Frobenius norm is given by the Moore-Penrose pseudoinverse \(\mat{X}^\dagger\) of $\mat X$ as
\begin{align}
    \mathbf{A} = \mathbf{Y}\mathbf{X}^\dagger.
\end{align}
The locally linear approximation can be then written as 
 $   \mat{Y} \approx \mat{A}\mat{X}.$

Now the DMD method aims to produce a low-rank eigen-decomposition of the matrix \(\mat{A}\).
Since the matrix \(\mathbf{A}\) is typically large the DMD algorithm proceeds in a numerically stable way as follows:
\begin{enumerate}
    \item The (reduced) singular value decomposition of $\mat{X}$ is carried out:
    \[
        \mat{X} = \mat{U}\Sigma\mat{V}^T,
    \]
    where $\mat U \in \R^{n, r}$, $\Sigma \in \R^{r,r}$ is diagonal, $\mat V \in \R^{m, r}$, and $r$ is the rank of $\mat X$.
    \item The matrix $\mat{A}$ is projected onto an $r$-dimensional subspace of $\R^n$ with orthonormal basis given by columns of $\mat U$:
    \begin{equation}\label{eq:DMD-Atilde}
        \tilde{\mat{A}}=\mat{U}^T\mat{A}\mat{U}=\mat{U}^T\mat{Y}\mat{V}\Sigma^{-1}.
    \end{equation}
    Here, the matrix $\tilde{\mat{A}} \in \R^{r,r}$ defines a low-dimensional linear model in the new coordinates:
    \[
        \tilde{\vec{x}}_{k+1}=\tilde{\mat{A}}\tilde{\vec{x}}_{k},
    \]
    where $\tilde{\vec x}_k = \mat U^T \vec x_k$ for all $k = 1,\dotsc,m$.
    \item The eigen-decomposition of $\tilde{\mat{A}}$ is performed as
     $   \tilde{\mat{A}}= \mat{W}\Lambda \mat W^*,$
    where $\Lambda$ is a diagonal matrix with eigenvalues, columns of $\mat{W}$ are the corresponding eigenvectors, and $\mat W^*$ is a conjugate transpose of $\mat W$.
    \item The final eigen-decomposition of $\mat{A}$ can be now reconstructed from $\mat{W}$ and $\Lambda$. The eigenvalues of $\mat{A}$ are given by $\Lambda$ and called the \emph{DMD eigenvalues}.
    The corresponding eigenvectors called the \emph{(exact) DMD modes} are the columns of a matrix $\Phi$ given by
     $   \Phi = \mat{Y}\mat{V}\Sigma^{-1}\mat{W}.$
\end{enumerate}

\begin{figure*}
    \centering
    \includegraphics[width=0.7\linewidth]{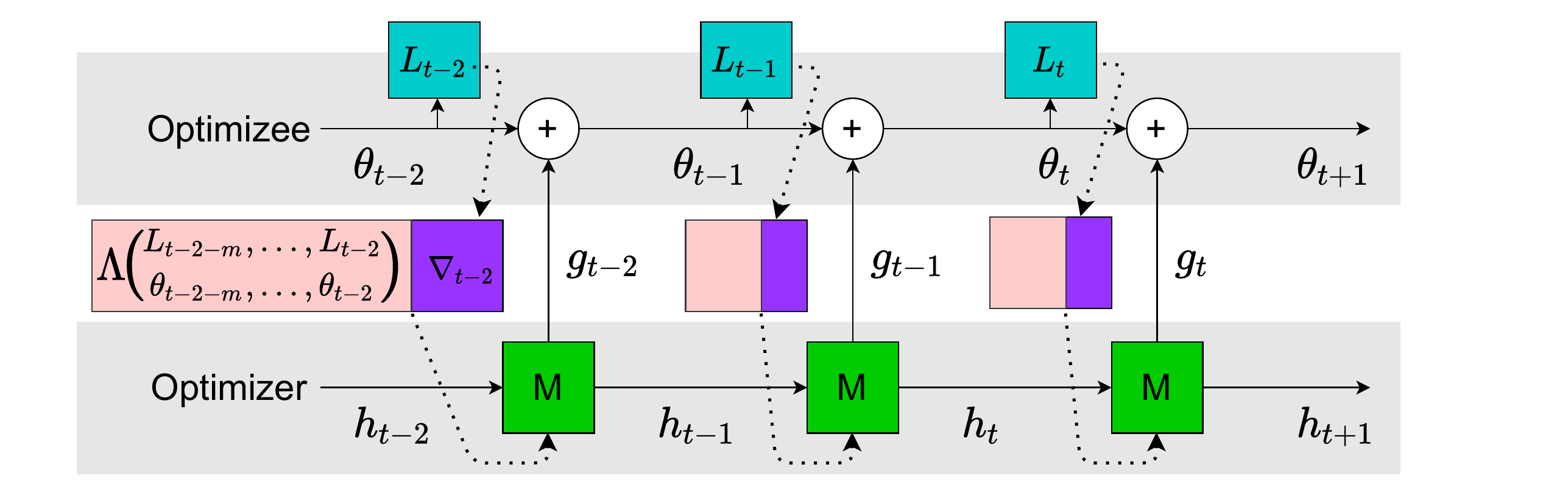}
    \caption{Graph used for computing the gradient of the optimizer in our L2O-DMD Method.}\label{fig:l2l-dmd}
\end{figure*}


The DMD modes describe spatial temporally coherent structures and the eigenvalues describe the dynamic behavior of the modes. The eigenvalues give us the information, whether the corresponding mode is exponentially growing, stable, or decaying, and whether the modes are oscillating.

It should be noted that there are many other versions of the DMD, 
see e.g. \citep{Kutz} for the description of some of them.

\section{Learning to Optimize with Dynamic Mode Decomposition}\label{sec:L2O-DMD}
The training of the parameters $\theta$ of the optimizee using an iterative learning algorithm can be understood as the evolution of a dynamical system. 
In the original L2O the parameters evolve in isolation and are not informed about the evolution of other parameters of the optimizee. 

It is natural to assume, that the performance can be improved by sharing some information about the evolution of all the parameters.
To achieve this, one could employ or learn some communication protocol between the optimizers similarly as e.g. in \citep{badger}. This can be very time-consuming and difficult to learn. 

Instead, we assume that informing each instance of the optimizer by dynamics of the overall optimization procedure might be helpful. The idea very vaguely follows the work of \citet{prokopenko}. The objective is that a dynamical system that is capable of producing novel methods must be self-referential and must be somehow able to encode the famous \emph{Liar paradox}. The Liar paradox for dynamical systems can be restated as “the system is not stable if and only if it can be shown to be stable”. 
Based on this we suggest an assumption that if the optimizer knows its dynamics and stability it could use that information to improve the efficiency of the learning procedure.

Let us propose the \emph{learning to optimize with dynamic mode decomposition} method (L2O-DMD) which uses DMD with reduced dimensionality to approximate the 
dynamics of the parameters and the loss function (on the fly during their evolution) and incorporates this information into the L2O procedure. 

We start by creating matrices $\mat X_t$ and $\mat Y_t$ of snapshots at each training step $t$ that store all the parameters of the optimizee as well as its losses in the last $m$ training steps:
\begin{align*}
    \mat X_t = 
    \begin{pmatrix}
        | &  & | \\
        \mathbf{\theta}_{t-m} & \cdots & \mathbf{\theta}_{t-1} \\
        | &  & |  \\
        L_{t-m} & & L_{t-1}
    \end{pmatrix},
    \mat Y_t = 
    \begin{pmatrix}
        | &  & | \\
        \mathbf{\theta}_{t-m +1} & \cdots & \mathbf{\theta}_{t} \\
        | &  & |  \\
        L_{t-m + 1} & & L_{t}
    \end{pmatrix},
\end{align*}
where $L_i = L(\theta_i)$ for all $i = t-m, \dotsc, t$.

The dimension of the exact DMD algorithm output is given by the rank of the matrix $\mat X_t$ that for a large number of parameters in $\theta$ typically equals $m$. In order to further control this dimension, we introduce a hyper-parameter $R \leq m$, called the DMD approximation rank, and use it in the second step of the DMD algorithm to reduce the output of the SVD where we assume the decreasing order of eigenvalues in $\Sigma_t$.
Hence, instead of matrices $\mat U_t, \Sigma_t, \mat V_t$ we use matrices $\tilde{\mat U}_t$ having the first $R$ columns of $\mat U_t$, $\tilde{\Sigma}_t$ diagonal having the first $R$ values of $\Sigma_t$, and $\tilde{\mat V}_t$ having the first $R$ rows of $\mat V_t$.
Thus, Equation \eqref{eq:DMD-Atilde} is changed to
\[
    \tilde{\mat{A}}_t= \tilde{\mat U}_t^T\mat{Y}_t\tilde{\mat V}_t\tilde{\Sigma}_t^{-1}.
\]
Then we proceed as usual. 
The resulting eigenvalues $\Lambda_t$ are fed as new features to each optimizer's instance of L2O.
For $t < m + 1$ where we don't have sufficient data for the DMD we set $\Lambda_t = \text{diag}(0)$.

The optimizer thus receives aggregate information about the dynamics of all the parameters and loss of the optimizee in the previous $m$ steps. As is indicated by the following experiments, these features are very strong and help the generalization of the L2O-DMD.

At every training step $t$ the optimizer $M$ calculates an update \eqref{eq:L2O-update} of parameters $\pvec\theta_t$ of the optimizee by $\pvec g_t$ based on the gradient $\nabla_{\pvec\theta} L(\pvec\theta_t)$, the DMD eigenvalues $\Lambda_t$, and its own hidden states $\pvec h_t$ as
\begin{equation}\label{eq:L2O-DMD-update}
    [\pvec g_t, \pvec h_{t + 1}] = M\big(\nabla_{\pvec\theta} L(\pvec\theta_t), \Lambda_t, \pvec h_t, \pvec\phi\big).
\end{equation}
The graph used for computing the gradient of the optimizer is presented in Figure \ref{fig:l2l-dmd}. In Figure \ref{fig:archb} we present the visualization of one optimization step of the L2O-DMD. 
Training of L2O-DMD is analogous to training of the L2O and is presented in Algorithm \ref{alg:l2o-dmd}. 
\begin{algorithm}
    \small
    \DontPrintSemicolon
    \SetAlgoLined
    \KwInput{Optimizee $L$ with parameters $\pvec\theta$, optimizer $M$ with parameters $\pvec\phi$ and hidden states $\pvec h$, training dataset $\mathscr{D}_{\text{train}}$}
    \KwParams{Number of training steps of the optimizee $T$, number of outer loop epochs $N$, unroll $u$, optimizer learning rate $\alpha$, optimizer loss weights $w_t$, DMD approximation rank $R$, number of snapshots $m$}
    \KwResult{Trained parameters $\pvec \phi$ of $M$}
    \BlankLine
    initialize $\pvec \phi$ of $M$\;
    \For{$i \leftarrow 1$ \KwTo $N$}
    {
       	initialize $\pvec \theta_1$ of $L$;\quad initialize $\pvec h_1$ of $M$\;
       	$\tau \leftarrow 1$;\quad $\Lambda_1 \leftarrow 0$ \;
       	\For{$t \leftarrow 1$ \KwTo $T$}
        {
            $L_\tau \leftarrow L(\pvec \theta_t)$\;
            \uIf{$t>m$}
            {
                $\mat X_t \leftarrow \big([\theta_{t-m}, L_{t-m}], \dotsc, [\theta_{t-1}, L_{t-1}]\big)$;\quad 
                $\mat Y_t \leftarrow \big([\theta_{t-m+1}, L_{t-m+1}], \dotsc, [\theta_{t}, L_{t}]\big)$\;
                $\mat U_t, \Sigma_t, \mat V_t \leftarrow \text{SVD}(\mat X_t)$\;
                \tcp{Reduce the dimension to $R$}
                $\tilde{\mat U}_t = \mat U_t[:, 1:R]$; \quad 
                $\tilde{\Sigma}_t = \Sigma_t[1:R, 1:R]$; \quad
                $\tilde{\mat V}_t = \mat V_t[:, 1:R]$\;
                $\tilde{\mat A}_t \leftarrow \tilde{\mat U}_t^T \mat Y_t \tilde{\mat V}_t \tilde{\Sigma}_t^{-1}$\;
                $\Lambda_t \leftarrow \text{EIG}(\tilde{\mat A}_t)$\quad \tcp{Compute DMD eigenvalues}
            }
            \Else{
                $\Lambda_t \leftarrow 0$\;
            }
            \tcp{Update optimizee}
            $\pvec g_{t}, \pvec h_{t+1} \leftarrow M\big(\nabla_{\pvec\theta} L(\pvec\theta_t), \Lambda_t, \pvec h_t, \pvec\phi\big)$\;
            $\pvec\theta_{t+1} \leftarrow \pvec\theta_t + g_t$\;
            \If{$\tau = u$}
            {   
                $\mathscr L(\pvec\phi) = \sum_{s = 1}^{u} w_s L_{s}$\;
                $\pvec\phi \leftarrow \pvec\phi - \alpha\nabla_{\pvec\phi} \mathscr L(\pvec \phi)$\quad \tcp{Update optimizer by SGD}
                $\tau \leftarrow 0$\;
            }
            $\tau \leftarrow \tau+1$\;
         }
    }
    \caption{Training of the L2O-DMD.}
    \label{alg:l2o-dmd}
\end{algorithm}

 After we described the algorithm, we want to explain the intuition behind the method. Why should the DMD eigenvalues matter? If a gradient of a particular parameter is small and the overall dynamics is fluctuating (eigenvalues close to 1), the optimizer could decide to do a much larger update than SGD would do based only on the gradient. Contrarily, if a particular LSTM receives information that the loss is low and the dynamics of the parameters is converging (eigenvalues < 1) but a particular gradient is still high, the optimizer can do a smaller update than SGD would. Increasing the rank $R$ can give the optimizer finer information about the dynamics while increasing the number of snapshots $m$ increases the history of the dynamic signal.

\section{Experiments}\label{sec:experiments}
In this section, we experimentally show that the proposed model has a highly improved performance over the original L2O. We follow the experimental setup from \citep{Andrychowicz}. The optimizer network $M$ is given by an LSTM recurrent neural network with 2 layers, each with 20 hidden units.
The results of the L2O-DMD are compared to the original L2O from Section \ref{section:l2o} and to the Adam optimizer with a learning rate of $0.01$. All the experiments were executed on one Tesla V100, the training takes around two hours.  
The methods are implemented in PyTorch \citep{NEURIPS2019_9015}\footnote{For the L2O we used implementation with MIT licence available at https://github.com/chenwydj/learning-to-learn-by-gradient-descent-by-gradient-descent.}.

\subsection{Training}\label{sec:training}
Both the L2O-DMD and the L2O optimizers are trained to minimize a cross-entropy of a small neural network $F$ classifying the MNIST dataset \citep{lecun-mnisthandwrittendigit-2010}. The neural network $F$ is a multi-layer perceptron with $1$ hidden layer with $8$ neurons. The sigmoid activation function is applied on the single neuron in the output layer. The optimizee $L$ 
is given by a cross-entropy loss of $F$ and its parameters $\theta$ are the parameters of $F$.

The values of the hyper-parameters used for training the L2O and the L2O-DMD are presented in Table \ref{tab:hyper}.
In the inner loop, the parameters of the optimizee are updated in $100$ training steps and the unroll is set to $20$.
In the outer loop, we ran $1000$ epochs. To improve the final result, we tested the performance of the current optimizer after every $20$th epoch in the following way.
We run training of $20$ new instances of the same optimizee on the testing dataset, with fixed parameters of the optimizer. Then we store the optimizer weights together with the performance measured by the optimizer loss \eqref{eq:optimizer-objective}, where the expectation is estimated by the mean.
After the outer loop finishes, we restore the weights of the best performing optimizer.

The parameters of the optimizer are optimized by the Adam method with a learning rate that was found out by testing multiple learning rates for the original L2O method. The best performing learning rate was taken and used also for the L2O-DMD. Thus, we did not tune the learning rate specifically for the L2O-DMD. 
This approach allows us to compare the methods more fairly and emphasize the DMD influence. 
\subsection{Testing Tasks}
Here we describe the tasks that are used for testing the performance of the already trained optimizers. In all scenarios, we run 1000 training steps of the optimizee, i.e. 1000 updates by the optimizer. All the presented results were trained as described in Section \ref{sec:training}.

\subsubsection{Image Classification}
We test the trained optimizer on various image classification tasks. The aim of the optimizer is to train small multilayer perceptron networks (MLP) with different depths, layer sizes, and activation functions on MNIST, Fashion MNIST \citep{xiao2017/online}, and CIFAR10 datasets \citep{cifar}. The optimizees are given by cross-entropy losses of the networks in all cases. 
The tasks denoted by \emph{MNIST-1L} and \emph{MNIST-2L} have $1$and $2$ hidden layers, respectively, with 20 neurons in each layer and sigmoid activation on the output. Please, note that the training was done on MLP with $8$ neurons. The optimizee training batch size is set to $128$. 
The task \emph{MNIST-Batch} is the same as MNIST-1L but with batch size 16. 
The lower batch size produces more noisy gradients and is very challenging for the original L2O algorithm. 
The task called \emph{MNIST-RELU} is the same as MNIST-1L but with the RELU activation instead of the sigmoid.
Finally, tasks called \emph{FashionMNIST-1L} and \emph{CIFAR10-1L} are analogous to MNIST-1L but on the FashionMNIST and CIFAR10 datasets, respectively. 
\subsection{Results}
Let us present and discuss the results of the testing tasks. Note that for all the scenarios the same optimizers trained as described in Section \ref{sec:training} are used. The performance is quantitatively evaluated using the optimizer loss \eqref{eq:optimizer-objective} where the expectation is estimated by the mean over 30 independent runs and the sum is taken over the last 10 training steps.
 More results including quadratic functions can found in Appendix \ref{quad}. Tests with different hyper-parameters of DMD are provided in 
\subsubsection{Image Classification}
\begin{figure}
    \centering
    \begin{subfigure}[b]{0.4\linewidth}
        \includegraphics[width=0.95\linewidth]{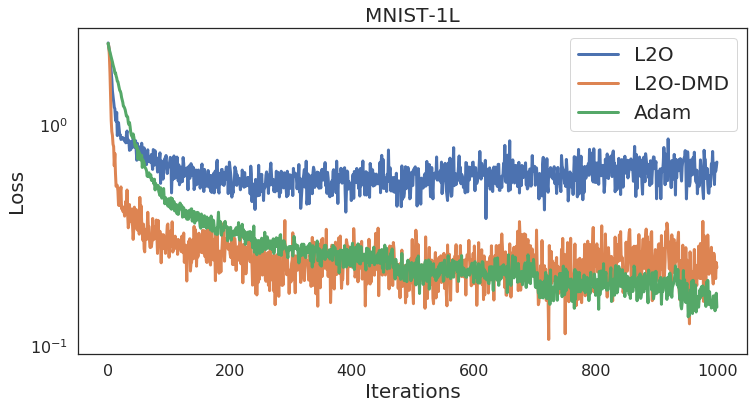}
        \caption{MLP with 1 hidden layer}\label{fig:mnist1}
    \end{subfigure}
    \begin{subfigure}[b]{0.4\linewidth}
        \includegraphics[width=0.95\linewidth]{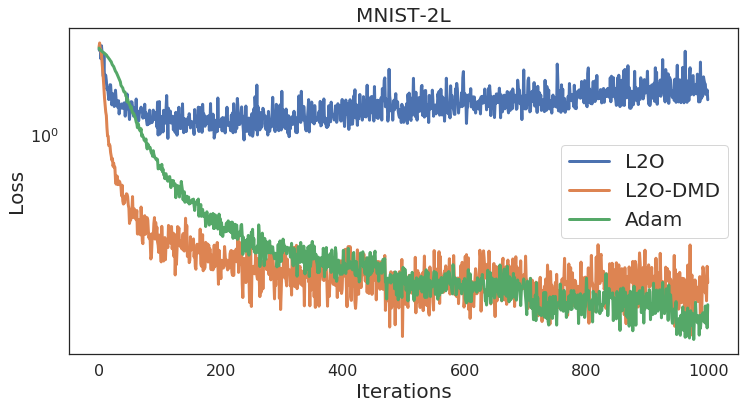}
        \caption{MLP with 2 hidden layers}\label{fig:mnist2}
    \end{subfigure}
    \begin{subfigure}[b]{0.4\linewidth}
        \includegraphics[width=0.95\linewidth]{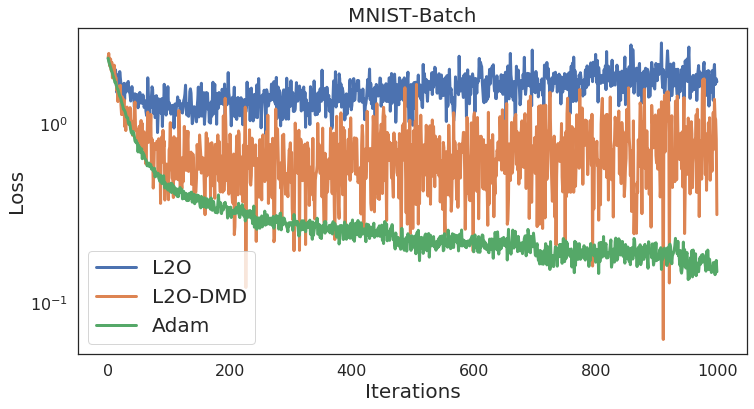}
        \caption{MLP with batch 16}\label{fig:batch}
    \end{subfigure}
    \begin{subfigure}[b]{0.4\linewidth}
        \includegraphics[width=0.95\linewidth]{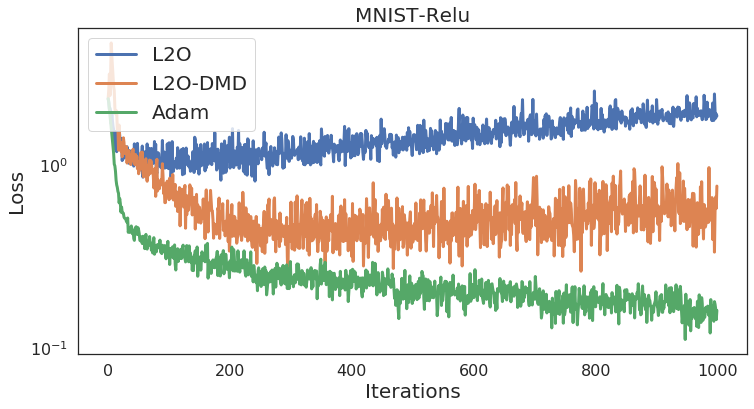}
        \caption{MLP with RELU}\label{fig:relu}
    \end{subfigure}
    \begin{subfigure}[b]{0.4\linewidth}
        \includegraphics[width=0.95\linewidth]{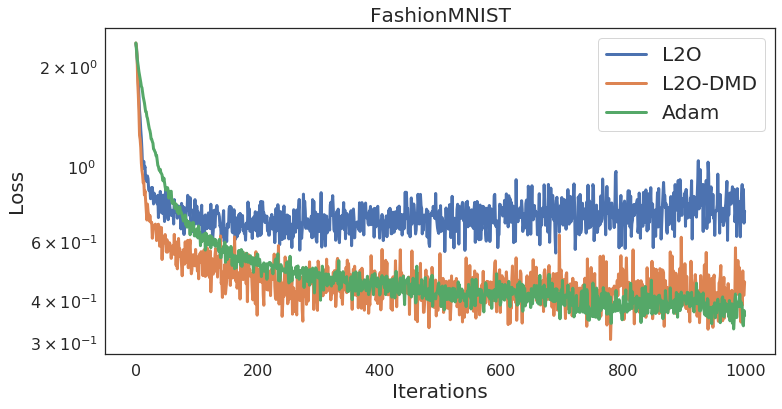}
        \caption{MLP training FashionMNIST}\label{fig:fashion}
    \end{subfigure}
    \begin{subfigure}[b]{0.4\linewidth}
        \includegraphics[width=0.95\linewidth]{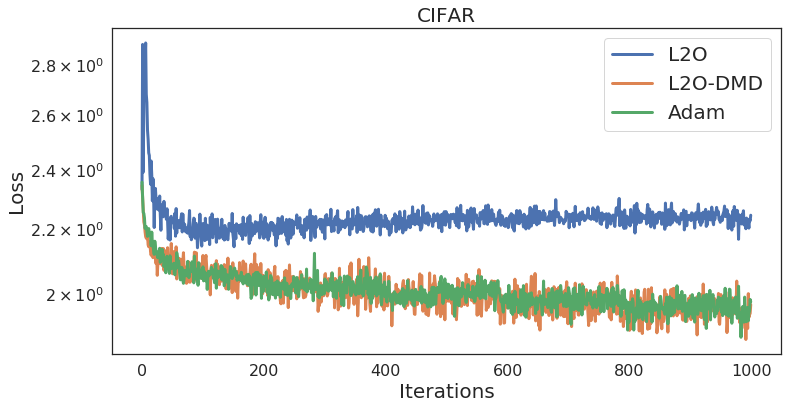}
       \caption{MLP training CIFAR10}\label{fig:cifar}
    \end{subfigure}
\caption{Comparing results on different tasks}
  \end{figure}

\begin{figure}
\centering
    \includegraphics[width=0.6\linewidth]{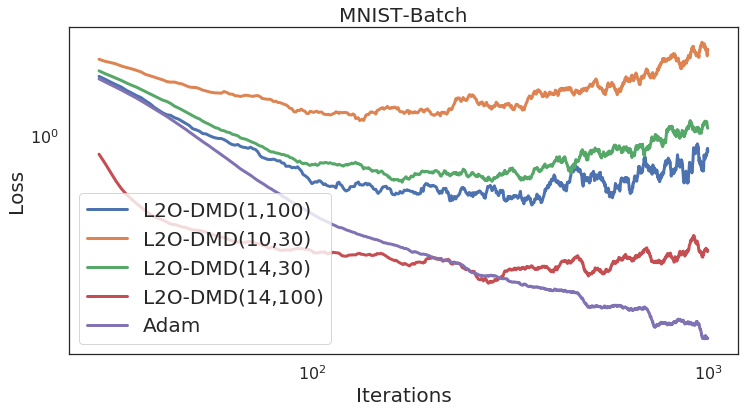}
    \caption{Mnist Batch task with different hyper-parameters of DMD, rolling mean of 30 iterations.}
    \label{fig:batch_comp}
\end{figure}

Results of the optimization of different image classification tasks are presented in Table \ref{tab:mnist}, we present here the averaged loss after $300$ iterations as in the original L2O paper. The L2O-DMD is always superior to the L2O but is outperformed by Adam on different batch size and on the RELU activation. 
In Figure \ref{fig:mnist1} we can see the learning curves on the same task that was used for training. From learning curves in Figure \ref{fig:mnist2} corresponding to the MNIST-2L task it follows that increasing the number of hidden layers to $2$ does not make any problems to both the L2O and the L2O-DMD, here the second is still superior to both the L2O and Adam. Decreasing the batch size shows to be difficult for all three methods, L2O fails while L2O-DMD struggles only in the later phase of the training, see Figure \ref{fig:batch}. Increasing the number of the snapshots included into DMD increases the performance of L2O-DMD as well as using more eigenvalues as can be seen in Figure \ref{fig:batch_comp}, here we use the parentheses to describe DMD hyperparameters $(R,m)$. 
As can be seen in Figure \ref{fig:relu}, both the L2O and the L2O-DMD have difficulties optimizing the network with the RELU activation function. The reason is that the different non-linearity of the activation function than what the optimizers were trained for (sigmoid) results in a very different loss landscape. The L2O-DMD is not as good as Adam but outperforms L2O significantly and thus demonstrates the improvement in generalization due to the use of the DMD. 
The improved generalization ability of the L2O-DMD can be also observed in Figures \ref{fig:fashion} and \ref{fig:cifar} corresponding to different datasets. The L2O-DMD optimizes significantly faster and achieves lower values of loss than both the L2O and Adam for the FashionMNIST dataset. On the more difficult CIFAR10 dataset where the L2O fails it is on a similar level with Adam.   
\begin{table}
    \centering
    \small
    \caption{Optimizer Loss for Image Classification Tasks.}\label{tab:mnist}
    \begin{tabular}{lccc}
      \toprule 
      \bfseries         & \bfseries L2O     & \bfseries L2O-DMD(1,100)  & \bfseries Adam\\
      MNIST-1L          & $0.55\pm 0.05$     & $\textbf{0.22}\pm0.46$        & $0.29\pm0.02$\\
      MNIST-2L          & $1.16\pm 0.10$     & $\textbf{0.26}\pm0.03$         & $0.26\pm0.02$\\
      MNIST-Batch       & $1.39\pm 0.22$    & $0.61\pm0.2$             & $\textbf{0.28}\pm0.02$\\
      MNIST-RELU        & $1.17\pm0.14$      & $0.44\pm0.08$                   & $\textbf{0.25}\pm0.03$\\
      FashionMNIST-1L   & $1.69\pm0.05$      & $\textbf{0.44}\pm0.05$        & $0.47\pm0.02$\\
      CIFAR10-1L        & $2.2\pm0.05$       & $\textbf{2.01}\pm0.02$        & $2.02\pm0.085$\\

      \bottomrule 
      
    \end{tabular}
\end{table}


\section{Limitations}\label{limitations}
The major limitation of the proposed method is similar to many other L2O methods, called a \emph{short-horizon bias} by \cite{wu2018understanding}. L2O methods often fail to converge for thousands of iterations, mostly because it is difficult to teach them to do so. L2O-DMD is better then L2O in short horizon and allow longer horizon without normalizing the gradients or other tricks that might constrain the network too much and result in method very similar to hand designed optimizer. We applied multiple ways how to increase the horizons and the results are described in Appendix \ref{longhor}. The proposed method currently improves the initial part of the training (which is by itself important task), we could include safeguarding by \cite{heaton2020safeguarded} in later stages of optimization. 
Scaling and computational overhead of DMD during inference can be another limiting factor. The inference overhead is around 5-10 percent. It can be reduced significantly by online DMD method that only updates the eigenvalues \citep{zhang2017online} and also by computing SVD and eigenvalues directly on GPUs. The scaling to larger problems could be a problem mostly because of the SVD used, it could be reduced by computing it layer-wise.  

\section{Conclusions and Future Steps}\label{sec:conclusion}
We have demonstrated that learning to optimize method successfully benefits from knowing the dynamics of the optimization procedure, extracted by the dynamic mode decomposition method. The undertaken experiments provide evidence that the optimizer trained with DMD generalizes much better than the original L2O method when applied to various architectures of neural networks, and distinct datasets.

In the future steps, we want to investigate various types of DMD to understand the effect of noisy gradients. Since we used DMD as a method to approximate information provided by the Koopman operator, we also want to explore more general approaches to how the Koopman operator could be used in learning to learn problems. We will also apply the method to optimization tasks where the gradients are not available.

\section{Broader Impact}\label{impact}
The proposed method contributes to the discovery of better performing learning algorithms. Improving L2O methods in general will lead to reduction of energy and labor costs of training ML models. We believe that such tools do not affect society in any harmful way. 

\bibliography{template.bib}

\begin{thebibliography}{28}
\providecommand{\natexlab}[1]{#1}
\providecommand{\url}[1]{\texttt{#1}}
\expandafter\ifx\csname urlstyle\endcsname\relax
  \providecommand{\doi}[1]{doi: #1}\else
  \providecommand{\doi}{doi: \begingroup \urlstyle{rm}\Url}\fi

\bibitem[Andrychowicz et~al.(2016)Andrychowicz, Denil, Colmenarejo, Hoffman,
  Pfau, Schaul, and Freitas]{Andrychowicz}
Marcin Andrychowicz, Misha Denil, Sergio~Gomez Colmenarejo, M.~W. Hoffman,
  D.~Pfau, T.~Schaul, and N.~D. Freitas.
\newblock Learning to learn by gradient descent by gradient descent.
\newblock In \emph{NIPS}, 2016.

\bibitem[Chen et~al.(2020)Chen, Zhang, Jingyang, Chang, Liu, Amini, and
  Wang]{ll2o}
Tianlong Chen, Weiyi Zhang, Zhou Jingyang, Shiyu Chang, Sijia Liu, Lisa Amini,
  and Zhangyang Wang.
\newblock Training stronger baselines for learning to optimize.
\newblock In \emph{Advances in Neural Information Processing Systems},
  volume~33, pages 7332--7343. Curran Associates, Inc., 2020.

\bibitem[Chen et~al.(2017)Chen, Hoffman, Colmenarejo, Denil, Lillicrap,
  Botvinick, and de~Freitas]{l2lw}
Yutian Chen, Matthew~W. Hoffman, Sergio~G{\'o}mez Colmenarejo, Misha Denil,
  Timothy~P. Lillicrap, Matt Botvinick, and Nando de~Freitas.
\newblock Learning to learn without gradient descent by gradient descent.
\newblock In \emph{Proceedings of the 34th International Conference on Machine
  Learning}, volume~70 of \emph{Proceedings of Machine Learning Research},
  pages 748--756, International Convention Centre, Sydney, Australia, 06--11
  Aug 2017. PMLR.

\bibitem[Dogra and Redman(2020)]{redman}
Akshunna~S. Dogra and William Redman.
\newblock Optimizing neural networks via koopman operator theory.
\newblock In \emph{Advances in Neural Information Processing Systems},
  volume~33, pages 2087--2097. Curran Associates, Inc., 2020.

\bibitem[Heaton et~al.(2020)Heaton, Chen, Wang, and Yin]{heaton2020safeguarded}
Howard Heaton, Xiaohan Chen, Zhangyang Wang, and Wotao Yin.
\newblock Safeguarded learned convex optimization, 2020.

\bibitem[Hochreiter and Schmidhuber(1997)]{LSTM}
S.~Hochreiter and J.~Schmidhuber.
\newblock Long short-term memory.
\newblock \emph{Neural Computation}, 9:\penalty0 1735--1780, 1997.

\bibitem[{Hospedales} et~al.(2020){Hospedales}, {Antoniou}, {Micaelli}, and
  {Storkey}]{review}
Timothy {Hospedales}, Antreas {Antoniou}, Paul {Micaelli}, and Amos {Storkey}.
\newblock {Meta-Learning in Neural Networks: A Survey}.
\newblock \emph{ArXiv}, abs/2004.05439, April 2020.

\bibitem[Kingma and Ba(2015)]{Adam}
Diederik~P. Kingma and Jimmy Ba.
\newblock Adam: A method for stochastic optimization.
\newblock \emph{CoRR}, abs/1412.6980, 2015.

\bibitem[Kirsch and Schmidhuber(2020)]{vsml}
Louis Kirsch and J.~Schmidhuber.
\newblock Meta learning backpropagation and improving it.
\newblock \emph{ArXiv}, abs/2012.14905, 2020.

\bibitem[Krizhevsky(2012)]{cifar}
Alex Krizhevsky.
\newblock Learning multiple layers of features from tiny images.
\newblock \emph{University of Toronto}, 05 2012.

\bibitem[Kutz et~al.(2017)Kutz, Brunton, Brunton, and Proctor]{Kutz}
Jose~Nathan Kutz, Steven~L. Brunton, Bingni~W. Brunton, and Joshua~L. Proctor.
\newblock \emph{Dynamic mode decomposition: data-driven modeling of complex
  systems}.
\newblock Society for Industrial and Applied Mathematics, 2017.

\bibitem[LeCun and Cortes(2010)]{lecun-mnisthandwrittendigit-2010}
Yann LeCun and Corinna Cortes.
\newblock {MNIST} handwritten digit database.
\newblock 2010.
\newblock URL \url{http://yann.lecun.com/exdb/mnist/}.

\bibitem[Lv et~al.(2017)Lv, Jiang, and Li]{rnnprop}
Kaifeng Lv, S.~Jiang, and J.~Li.
\newblock Learning gradient descent: Better generalization and longer horizons.
\newblock In \emph{ICML}, 2017.

\bibitem[{Manojlovi{\'c}} et~al.(2020){Manojlovi{\'c}}, {Fonoberova}, {Mohr},
  {Andrej{\v{c}}uk}, {Drma{\v{c}}}, {Kevrekidis}, and {Mezi{\'c}}]{Mezic}
Iva {Manojlovi{\'c}}, Maria {Fonoberova}, Ryan {Mohr}, Aleksandr
  {Andrej{\v{c}}uk}, Zlatko {Drma{\v{c}}}, Yannis {Kevrekidis}, and Igor
  {Mezi{\'c}}.
\newblock {Applications of Koopman Mode Analysis to Neural Networks}.
\newblock \emph{arXiv}, abs/2006.11765, June 2020.

\bibitem[Metz et~al.(2020)Metz, Maheswaranathan, Freeman, Poole, and
  Sohl-Dickstein]{metz}
Luke Metz, Niru Maheswaranathan, C.~D. Freeman, Ben Poole, and Jascha
  Sohl-Dickstein.
\newblock Tasks, stability, architecture, and compute: Training more effective
  learned optimizers, and using them to train themselves.
\newblock \emph{ArXiv}, abs/2009.11243, 2020.

\bibitem[Metz et~al.(2021)Metz, Freeman, Maheswaranathan, and
  Sohl-Dickstein]{metz2}
Luke Metz, C.~Daniel Freeman, Niru Maheswaranathan, and Jascha Sohl-Dickstein.
\newblock Training learned optimizers with randomly initialized learned
  optimizers.
\newblock \emph{ArXiv}, abs/2101.07367, 2021.

\bibitem[Paszke et~al.(2019)Paszke, Gross, Massa, Lerer, Bradbury, Chanan,
  Killeen, Lin, Gimelshein, Antiga, Desmaison, Kopf, Yang, DeVito, Raison,
  Tejani, Chilamkurthy, Steiner, Fang, Bai, and Chintala]{NEURIPS2019_9015}
Adam Paszke, Sam Gross, Francisco Massa, Adam Lerer, James Bradbury, Gregory
  Chanan, Trevor Killeen, Zeming Lin, Natalia Gimelshein, Luca Antiga, Alban
  Desmaison, Andreas Kopf, Edward Yang, Zachary DeVito, Martin Raison, Alykhan
  Tejani, Sasank Chilamkurthy, Benoit Steiner, Lu~Fang, Junjie Bai, and Soumith
  Chintala.
\newblock Pytorch: An imperative style, high-performance deep learning library.
\newblock In H.~Wallach, H.~Larochelle, A.~Beygelzimer, F.~d\textquotesingle
  Alch\'{e}-Buc, E.~Fox, and R.~Garnett, editors, \emph{Advances in Neural
  Information Processing Systems 32}, pages 8024--8035. Curran Associates,
  Inc., 2019.
\newblock URL
  \url{http://papers.neurips.cc/paper/9015-pytorch-an-imperative-style-high-performance-deep-learning-library.pdf}.

\bibitem[Prokopenko et~al.(2019)Prokopenko, Harr{\'e}, Lizier, Boschetti,
  Peppas, and Kauffman]{prokopenko}
M.~Prokopenko, M.~Harr{\'e}, J.~Lizier, F.~Boschetti, P.~Peppas, and
  S.~Kauffman.
\newblock Self-referential basis of undecidable dynamics: from the liar paradox
  and the halting problem to the edge of chaos.
\newblock \emph{Physics of life reviews}, 2019.

\bibitem[Rosa et~al.(2019)Rosa, Afanasjeva, Andersson, Davidson, Guttenberg,
  Hlubuček, Poliak, Vítku, and Feyereisl]{badger}
Marek Rosa, Olga Afanasjeva, Simon Andersson, Joseph Davidson, Nicholas
  Guttenberg, Petr Hlubuček, Martin Poliak, Jaroslav Vítku, and Jan
  Feyereisl.
\newblock Badger: Learning to (learn [learning algorithms] through multi-agent
  communication).
\newblock \emph{ArXiv}, abs/1912.01513, 2019.

\bibitem[Schmid(2008)]{Schmid}
P.~Schmid.
\newblock Dynamic mode decomposition of numerical and experimental data.
\newblock \emph{Journal of Fluid Mechanics}, 656:\penalty0 5--28, 2008.

\bibitem[Schmidhuber(2020)]{schmidhuber}
Juergen Schmidhuber, 2020.
\newblock URL \url{https://people.idsia.ch/~juergen/metalearning.html}.

\bibitem[Tu et~al.(2014)Tu, Rowley, Luchtenburg, Brunton, and Kutz]{tu2014}
Jonathan~H. Tu, Clarence~W. Rowley, Dirk~M. Luchtenburg, Steven~L. Brunton, and
  J.~Nathan Kutz.
\newblock On dynamic mode decomposition: Theory and applications.
\newblock \emph{Journal of Computational Dynamics}, 1\penalty0 (2):\penalty0
  391--421, Dec 2014.
\newblock \doi{10.3934/jcd.2014.1.391}.

\bibitem[Wichrowska et~al.(2017)Wichrowska, Maheswaranathan, Hoffman,
  Colmenarejo, Denil, de~Freitas, and Sohl-Dickstein]{l2o-scale}
Olga Wichrowska, Niru Maheswaranathan, Matthew~W. Hoffman, Sergio~G{\'o}mez
  Colmenarejo, Misha Denil, Nando de~Freitas, and Jascha Sohl-Dickstein.
\newblock Learned optimizers that scale and generalize.
\newblock In \emph{Proceedings of the 34th International Conference on Machine
  Learning}, volume~70 of \emph{Proceedings of Machine Learning Research},
  pages 3751--3760, International Convention Centre, Sydney, Australia, 06--11
  Aug 2017. PMLR.

\bibitem[Williams et~al.(2015)Williams, Kevrekidis, and Rowley]{EDMD}
Matthew~O. Williams, Ioannis~G. Kevrekidis, and Clarence~W. Rowley.
\newblock A data--driven approximation of the koopman operator: Extending
  dynamic mode decomposition.
\newblock \emph{Journal of Nonlinear Science}, 25\penalty0 (6):\penalty0
  1307--1346, Dec 2015.

\bibitem[Wu et~al.(2018)Wu, Ren, Liao, and Grosse]{wu2018understanding}
Yuhuai Wu, Mengye Ren, Renjie Liao, and Roger Grosse.
\newblock Understanding short-horizon bias in stochastic meta-optimization,
  2018.

\bibitem[Xiao et~al.(2017)Xiao, Rasul, and Vollgraf]{xiao2017/online}
Han Xiao, Kashif Rasul, and Roland Vollgraf.
\newblock Fashion-mnist: a novel image dataset for benchmarking machine
  learning algorithms, 2017.

\bibitem[Zhang et~al.(2017)Zhang, Rowley, Deem, and
  Cattafesta]{zhang2017online}
Hao Zhang, Clarence~W. Rowley, Eric~A. Deem, and Louis~N. Cattafesta.
\newblock Online dynamic mode decomposition for time-varying systems, 2017.

\bibitem[Zhou et~al.(2017)Zhou, Li, and Zare]{chemopt}
Zhenpeng Zhou, Xiaocheng Li, and Richard~N. Zare.
\newblock Optimizing chemical reactions with deep reinforcement learning.
\newblock \emph{ACS central science}, 3\penalty0 (12):\penalty0 1337--1344, Dec
  2017.

\end{thebibliography}

%
%
%
%
%
%
%
%
%
%
%
%
%
%
%
%
%
%


\end{document}